\documentclass{bmvc2k}

\title{HyperVision: A Channel-Adaptive Ground-Based Hyperspectral Vision Pre-trained Backbone}

\addauthor{Guanyiman Fu}{guanyiman.fu@griffith.edu.au}{1}
\addauthor{Jingtao Li$^*$}{jl2692@cam.ac.uk}{2}
\addauthor{Zihang Cheng}{chengzihang@njust.edu.cn}{3}
\addauthor{Zhuanfeng Li}{lizhuanfeng@hytc.edu.cn}{4,5}
\addauthor{Diqi Chen}{D.Chen1@massey.ac.nz}{6}
\addauthor{Yan Xu}{xuyan@njust.edu.cn}{3}
\addauthor{Xiangyu Liu}{xiangyu.liu@griffith.edu.au}{1}
\addauthor{Fengchao Xiong}{fcxiong@njust.edu.cn}{3}
\addauthor{Jianfeng Lu$^*$}{lujf@njust.edu.cn}{3}
\addauthor{Chengrong Chen}{c.chen@griffith.edu.au}{1}
\addauthor{Jun Zhou}{jun.zhou@griffith.edu.au}{7}

\addinstitution{
Australian Rivers Institute and School of Environment and Science,\\
Griffith University, Australia
}
\addinstitution{
Department of Computer Science and Technology,\\
University of Cambridge, UK
}
\addinstitution{
School of Computer Science and Engineering,\\
Nanjing University of Science and Technology, China
}
\addinstitution{
 School of Computer Science and Technology,\\
 Huaiyin Normal University, China
}
\addinstitution{
State Key Lab. of Novel Software Technology,\\
Nanjing University, China
}
\addinstitution{
School of Agriculture and Environment,\\
Massey University, New Zealand
}
\addinstitution{
School of Information and Communication Technology,\\
Griffith University, Australia\\ \vspace{0.3cm}
 $^*$ Corresponding authors.
}

\runninghead{Fu et al.}{HyperVision}

\usepackage{amsmath}
\usepackage{amssymb}
\usepackage{booktabs}
\usepackage{algorithm}
\usepackage{algorithmic}
\usepackage{xcolor}
\usepackage{multirow}
\usepackage{array}
\usepackage{url}

\def\eg{\emph{e.g}\bmvaOneDot}

\begin{document}

\maketitle

\vspace{-0.7cm}
\begin{abstract}
    \vspace{-0.1cm}
    While hyperspectral imaging provides rich spatial-spectral information across hundreds of narrow wavelength bands for precise material identification, ground-based hyperspectral pre-trained backbones remain absent, constrained by varying spectral configurations across sensors, limited annotations and heterogeneous labeling schemes, and the limited scale and scene diversity of existing datasets. To address these challenges and enable universal perception, we propose HyperVision, the first ground-based hyperspectral pre-trained backbone. First, to handle varying spectral configurations, HyperVision adopts a channel-adaptive dynamic embedding mechanism to map heterogeneous inputs into a unified token space. Second, we develop an unsupervised representation learning framework. Specifically, to address limited annotations and heterogeneous labeling schemes, a multi-source pseudo-labeling method is introduced to fuse spatial structures from SAM2 and fine-grained spectral material information from HyperFree. Furthermore, to enrich scene diversity and compensate for limited dataset scale, a cross-modal knowledge distillation mechanism is utilized to transfer rich semantic representations from a pre-trained RGB vision model to our backbone. Pre-trained on a collection of 15k images from 26 diverse ground-based datasets, HyperVision demonstrates exceptional generalization. Requiring only efficient head-only adaptation without adjusting backbone parameters, it outperforms state-of-the-art task-specific methods across three downstream tasks under varying sensor configurations, yielding up to a 16.3\% relative improvement in hyperspectral semantic segmentation $\mathrm{Acc}_{\mathrm{M}}$, a 2.1\% relative gain in object tracking AUC, and a 35.5\% reduction in salient object detection MAE. The source code and pre-trained models are available at \url{https://github.com/lronkitty/HyperVision}.
\end{abstract}
\vspace{-0.1cm}


\vspace{-0.5cm}
\section{Introduction}
\vspace{-0.2cm}
\label{sec:intro}

Hyperspectral imaging provides rich spatial-spectral information by capturing scene reflect-ance across hundreds of narrow wavelength bands~\cite{Warnke2019100tech}. Unlike conventional RGB cameras, which compress spectral reflectances into three broad visible bands, hyperspectral images (HSIs) capture the intrinsic physical properties of materials. This capability enables precise material identification and robust perception even when objects are visually similar or intentionally camouflaged. Consequently, hyperspectral imaging has driven critical advancements in diverse ground-based vision applications, spanning from autonomous driving~\cite{hsidrive20} and salient object detection~\cite{hsodbitv2} to complex object tracking~\cite{mht} and camouflage detection~\cite{Hossain2023Camouflage}.

\begin{figure}[b]
    \centering
    \vspace{-0.5cm}
    \includegraphics[width=0.58\linewidth]{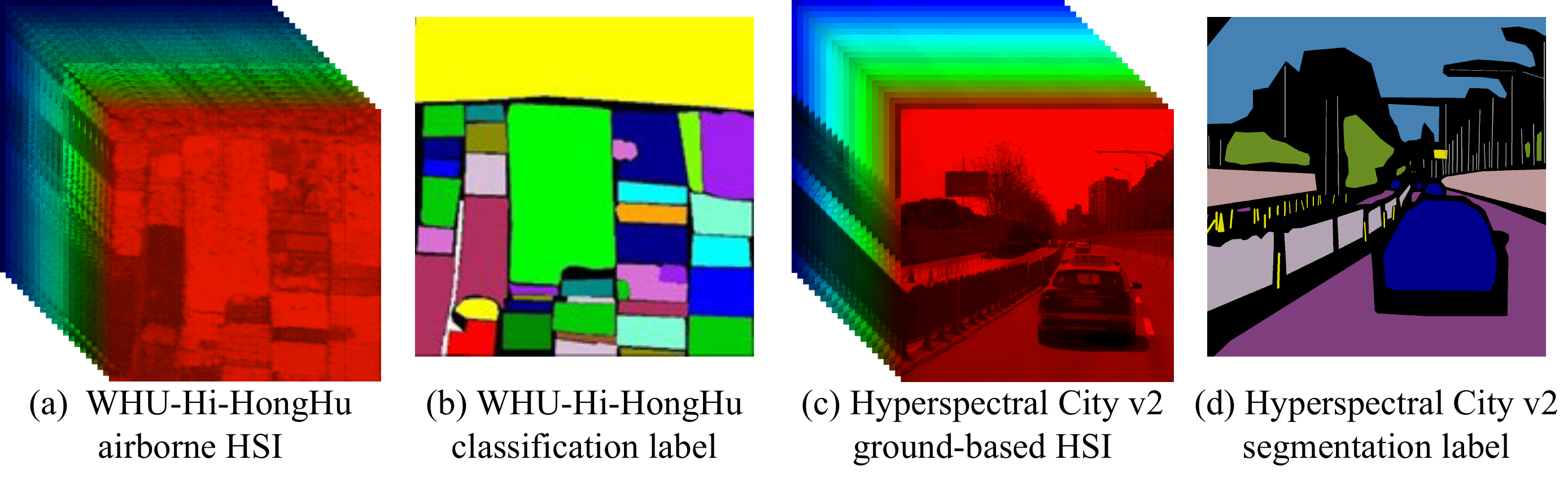}
    \vspace{-0.3cm}
    \caption{Comparison of airborne and ground-based HSI modeling.}\label{fig:airborne-ground}
    \vspace{-0.5cm}
\end{figure}

Recently, RGB vision pre-trained models have achieved remarkable success in general vision tasks. Driven by massive, homogenized datasets containing billions of images, models such as CLIP~\cite{radford2021learning}, MAE~\cite{he2022masked}, the SAM family~\cite{sam2, sam3_2025}, and the DINO series~\cite{oquab2024dinov2, simeoni2025dinov3} have demonstrated exceptional zero-shot generalization and robust feature representation capabilities. These models have revolutionized standard computer vision paradigms, shifting the field from task-specific training to prompt-driven and lightweight adaptation architectures.

Simultaneously, the abundance of standardized airborne imaging data has spurred the development of domain-specific pre-trained backbones for airborne HSIs, such as SpectralGPT~\cite{hong2024spectralgpt} and HyperFree~\cite{Li2025hyperfree}. However, the highly promising domain of ground-based hyperspectral analysis still lacks similar unified modeling paradigms. As illustrated in Figure~\ref{fig:airborne-ground}, airborne hyperspectral imaging, such as HongHu~\cite{whuhi}, typically focuses on large-scale terrain mapping and patch-level material classification from an orthographic perspective. In contrast, ground-based HSIs, such as in Hyperspectral City v2~\cite{hyperspectralcityv2}, require models to simultaneously understand complex spatial structures, high-level semantics, and fine-grained material compositions. This demands dense semantic segmentation to parse complex spatial topologies, perspective distortions, and distinct object boundaries within an egocentric view~\cite{hs3bench}. Consequently, due to this fundamental divergence in spatial complexity and task formulation, pre-trained backbones tailored for airborne scenarios are not optimally suited for direct application to ground-level perception.

\begin{figure}[t!]
    \centering
    \vspace{-0.1cm}
    \includegraphics[width=0.7\linewidth]{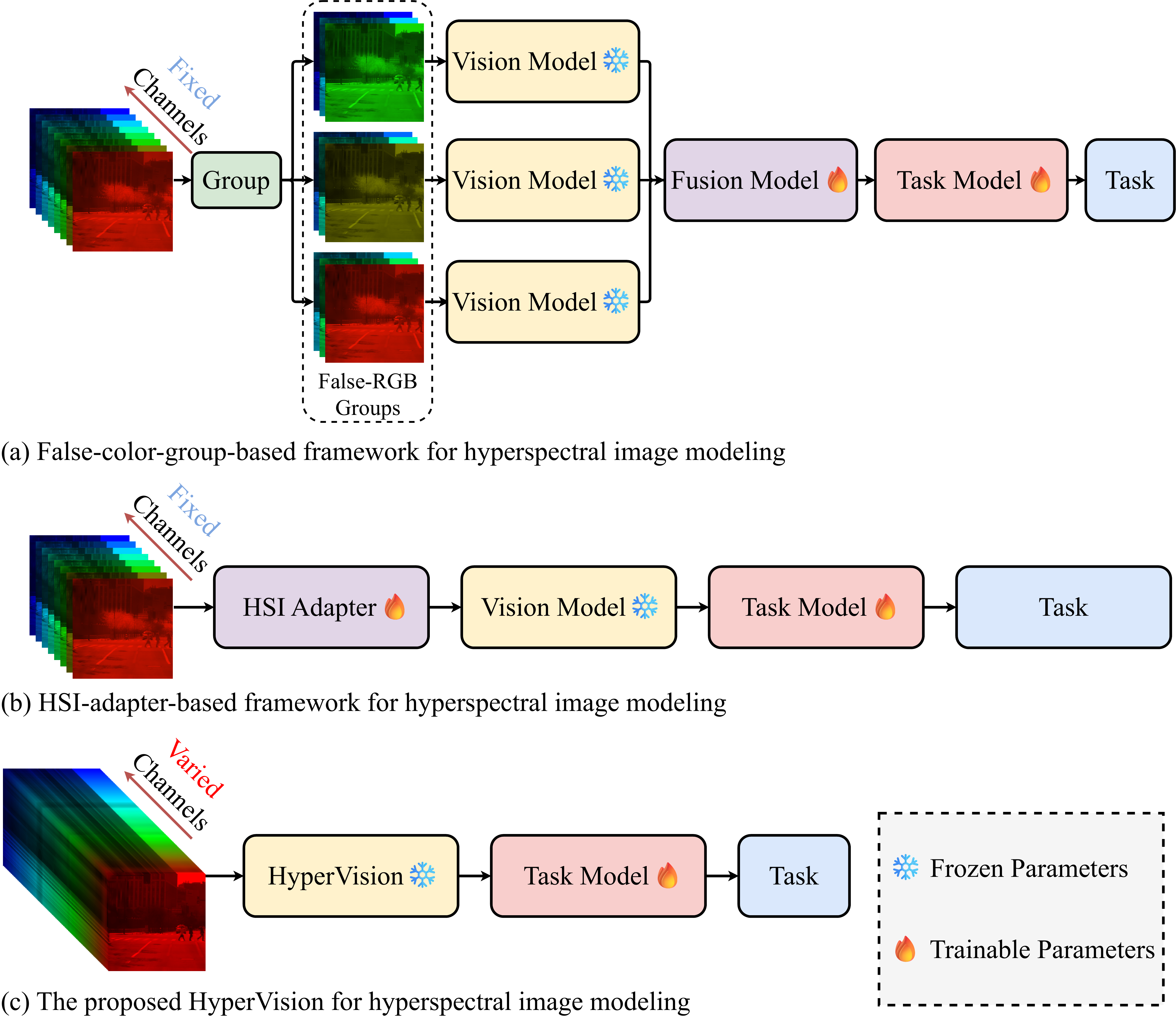}
    \vspace{-0.35cm}
    \caption{Comparison of existing pre-trained HSI modeling frameworks for downstream tasks with the proposed HyperVision framework.}\label{fig:hsis_processing}
    \vspace{-0.45cm}
\end{figure}

In the absence of dedicated pre-trained backbones for the ground-based scenarios, current methods usually adapt pretrained RGB vision models. As shown in Figure~\ref{fig:hsis_processing}(a) and (b), these approaches accommodate hyperspectral inputs by either grouping bands into multiple false-color images for separate processing and fusion~\cite{Gao2023CBFFNet,Li2023DENet}, or by employing specialized prompt adapters~\cite{hurtado2025hyperspectraladapter}. While these methods reuse existing visual priors to reduce the need for annotated data, their performance remains limited. Specifically, they rely on fixed input channel dimensions and RGB-centric architectures, which prevent them from fully exploiting dense spectral correlations.

Building a ground-based hyperspectral pre-trained backbone capable of extracting universal features directly from hyperspectral data would fundamentally resolve these limitations. However, developing such a model remains a challenge hindered by three major bottlenecks:

1) Hardware inconsistency: Unlike RGB cameras, ground-based HSI sensors exhibit significant variations in their spectral configurations, with channel counts ranging from 15 to 420 bands and wavelength coverage spanning from 377 nm to 1700 nm. This creates a ``hardware wall'' that prevents aggregating fragmented public datasets into a unified pre-training dataset collection.

2) Scarcity and inconsistency of labels: The HSI domain severely lacks the abundant weak supervision signals found in RGB images. Acquiring high-quality annotations is prohibitively expensive, and existing datasets exhibit significant label inconsistency across different tasks, which prevents joint training.

3) Scene sparsity: Due to complex acquisition constraints, existing ground-based HSI datasets are limited in scale and environmental diversity, making models susceptible to overfitting.

To address these limitations, we introduce HyperVision, the first pre-trained backbone for ground-based hyperspectral perception. To overcome the ``hardware wall'' of heterogeneous sensors, HyperVision employs a unified token space via a channel-adaptive mechanism~\cite{Li2025hyperfree}. By conceptualizing wavelengths as lookup keys, it dynamically retrieves and assembles network weights from a learned dictionary. This enables the network to accommodate arbitrary spectral inputs without modifying or retraining the backbone for different spectral configurations, ensuring varied channel compositions are encoded into a unified representation space.

Furthermore, to address limited, incompatible annotations and restricted scene diversity, we assemble a pre-training dataset collection comprising 15k HSIs across 26 diverse public datasets, and design an unsupervised representation learning framework. This framework consists of two core components: pseudo-label generation and knowledge distillation. First, to generate robust pseudo-targets, we propose a multi-source unsupervised pseudo-label generation strategy that fuses semantic representations derived from complex spatial structures via SAM2~\cite{sam2} and fine-grained spectral material information from the airborne model HyperFree~\cite{Li2025hyperfree}. Second, to further mitigate the issue of limited scene diversity and enrich the model's target-understanding capacity, we introduce a cross-modal knowledge distillation mechanism. By transferring rich semantic representations from pre-trained RGB vision models to our hyperspectral backbone, it acquires comprehensive scene understanding capabilities without requiring manual labels. Ultimately, HyperVision is pre-trained as a universal hyperspectral backbone that decouples downstream tasks from hardware constraints and achieves powerful generalization via parameter-efficient head-only adaptation, as shown in Figure~\ref{fig:hsis_processing}(c).

Our contributions can be summarized as follows:
\begin{itemize}
    \vspace{-0.1cm}
    \item We propose HyperVision, the first pre-trained backbone dedicated to ground-based hyperspectral perception. By adopting a channel-adaptive mechanism~\cite{Li2025hyperfree}, it integrates heterogeneous hyperspectral inputs into a unified token space, addressing the hardware variability of sensor configurations.
          \vspace{-0.1cm}

    \item We develop a multi-source pseudo-labeling strategy that leverages semantic representations from SAM2's~\cite{sam2} spatial structures and HyperFree's~\cite{Li2025hyperfree} fine-grained spectral material information. This strategy provides high-quality pseudo-labels for network training, effectively addressing label scarcity and cross-dataset annotation heterogeneity.
          \vspace{-0.1cm}

    \item We further introduce a cross-modal knowledge distillation strategy to transfer general visual priors from a pre-trained RGB vision model, enhancing target understanding and compensating for the limited scene diversity in hyperspectral data.
          \vspace{-0.1cm}

    \item We integrate the proposed pseudo-labeling and distillation strategies into a unified unsupervised representation learning framework. Pre-trained on a collection of 15k images from 26 diverse datasets, HyperVision demonstrates strong generalization, achieving state-of-the-art performance across hyperspectral semantic segmentation, object tracking, and salient object detection via efficient head-only adaptation on a frozen backbone.
\end{itemize}

\vspace{-0.7cm}
\section{Related Work}
\vspace{-0.2cm}

\subsection{Deep Learning for Hyperspectral Vision}
\vspace{-0.1cm}

Ground-based hyperspectral imaging extends beyond traditional airborne hyperspectral imaging to advanced tasks like autonomous driving~\cite{hsidrive20}, object tracking~\cite{mht}, and salient object detection~\cite{hsodbitv2}. Unlike RGB methods, deep learning models for HSIs exploit dense spectral correlations to capture intrinsic material properties. For instance, SEE-Net~\cite{Li2023DENet} learns band importance via a spectral self-expressive module to dynamically ensemble predictions for robust tracking. Similarly, SMN~\cite{liu2024smn} fuses low-frequency spectral saliency and high-frequency edges via mixed-frequency attention to accurately detect objects in challenging scenes. However, achieving further breakthroughs requires the powerful generalization of pre-trained backbones, whose development is severely bottlenecked by hardware fragmentation, label scarcity and inconsistency, and limited dataset diversity.

\vspace{-0.35cm}
\subsection{Vision Pre-trained Models}
\vspace{-0.15cm}
The rapid evolution of vision pre-trained models has revolutionized general-purpose perception in the computer vision domain. Trained on massive datasets, models like MAE~\cite{he2022masked} and the DINO series~\cite{oquab2024dinov2, simeoni2025dinov3} extract robust visual semantics that transfer across tasks. Furthermore, architectures like SAM2~\cite{sam2} introduced cross-task adaptability through prompt-driven, zero-shot spatial segmentation. Progressing significantly further, SAM3~\cite{sam3_2025} marks a paradigm shift from interactive spatial prompting to automated concept-level perception, achieving semantic-level understanding. Concurrently, deep learning methodologies have been extensively explored in large-scale airborne hyperspectral imaging, yielding robust domain-specific pre-trained backbones like SpectralGPT~\cite{hong2024spectralgpt} and HyperFree~\cite{Li2025hyperfree}. However, compared to these advanced vision and airborne pre-trained backbones, there remains a critical lack of dedicated pre-trained models for current ground-based hyperspectral vision tasks.

\vspace{-0.35cm}
\subsection{Pre-trained Backbones in Hyperspectral Vision}
\vspace{-0.15cm}
Initial efforts to introduce pre-trained models into the HSI domain primarily adapted pretrained RGB vision backbones. Some approaches split hyperspectral bands into three-channel false-color representations to directly feed into off-the-shelf RGB vision pre-trained models~\cite{Gao2023CBFFNet,Li2023DENet}, or utilize cross-modality feature modulation to align RGB and HSI distributions~\cite{Li2023RGB}. Other methods employ parameter-efficient tuning on RGB vision pre-trained backbones, such as the HSI-Adapter~\cite{hurtado2025hyperspectraladapter} and visual prompting~\cite{Zhu_2023_CVPR}. While these strategies effectively reuse strong visual priors, they strictly assume fixed input channel dimensions and rely on RGB-centric architectures. Consequently, they fail to fully exploit the dense spectral correlations inherent in hyperspectral data and are unable to accommodate the widely varying spectral bands acquired by different sensors.

In summary, establishing a dedicated ground-based hyperspectral pre-trained backbone is critical for unlocking the full potential of hyperspectral imaging in close-range perception. However, the development of such a universal model requires overcoming three critical bottlenecks: 1) hardware fragmentation across diverse HSI sensors; 2) the scarcity and inconsistency of labels; and 3) the limited scene diversity in existing HSI datasets. To overcome these, HyperVision systematically integrates a channel-adaptive mechanism~\cite{Li2025hyperfree} to handle heterogeneous spectral inputs, a multi-source pseudo-labeling paradigm bridging SAM2 and HyperFree to overcome annotation scarcity and heterogeneous labeling schemes, and cross-modal distillation from a robust RGB vision pre-trained model to compensate for scene sparsity.

\vspace{-0.45cm}
\section{HyperVision Backbone}
\vspace{-0.2cm}

\begin{figure*}[tbp]
    \centering
    \vspace{-0.1cm}
    \includegraphics[width=1\linewidth]{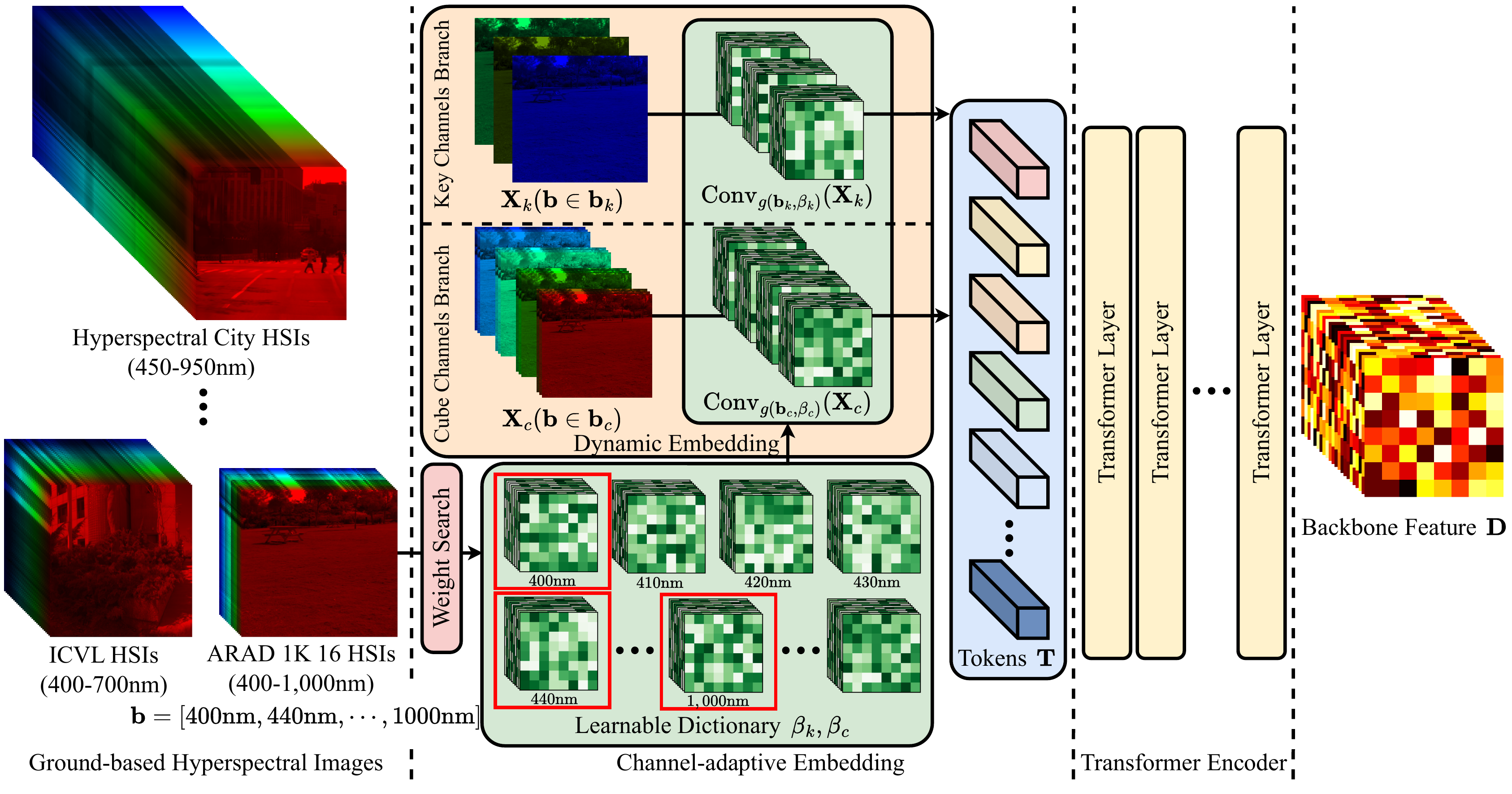}
    \vspace{-0.6cm}
    \caption{The architecture of HyperVision.}
    \label{fig:hypervision_arch}
    \vspace{-0.5cm}
\end{figure*}

HyperVision is a pre-trained backbone for ground-based hyperspectral perception, as shown in Figure~\ref{fig:hypervision_arch}. It adopts a channel-adaptive spectral modeling framework from HyperFree~\cite{Li2025hyperfree} to overcome the limitations of fixed-channel RGB architectures. Unlike existing methods that rely on specialized prompt-adapters or group hyperspectral bands into false-color subsets to fit standard vision models, the channel-adaptive mechanism treats individual wavelengths as lookup keys to dynamically assemble layer weights from a learned dictionary. This transforms hyperspectral inputs with arbitrary channel configurations into a unified, fixed-length token sequence, eliminating the need for task-specific retraining. Consequently, it enables efficient head-only adaptation to diverse perception tasks, significantly reducing training overhead while maintaining the powerful generalization required of a universal pre-trained backbone.

Specifically, following HyperFree~\cite{Li2025hyperfree}, we define the weight dictionary over the target wavelength range by assigning an index to every 10 nm interval of the spectrum. For an input HSI $\mathbf{X} \in \mathbb{R}^{h \times w \times c}$ with wavelength list $\mathbf{b} = [b_1, b_2, \ldots, b_c]$ and target token dimension $j$, we retrieve and concatenate the corresponding parameter tensors to dynamically form the embedding weights. Considering that the importance of varying wavebands differs in hyperspectral tasks~\cite{Li2023DENet, wang2020fast}, we extend the embedding stage with a two-branch design that processes dynamically assembled layers in parallel. Specifically, $\mathbf{X}$ is patchified in step $p$ and split into a key-channel component $\mathbf{X}_k$ and an intermediate-cube component $\mathbf{X}_c$. Denoting the dictionary-based weight construction as $g(\mathbf{b}, \beta)$, we use two separate dictionaries $\beta_k$ and $\beta_c$ to process the branches in parallel. The final visual tokens $\mathbf{T} \in \mathbb{R}^{\frac{h}{p} \times \frac{w}{p} \times j}$, are obtained by the element-wise summation of the convolution outputs of the two branches:
\begin{equation}
    \vspace{-0.05cm}
    \mathbf{T} = \text{Conv}_{g(\mathbf{b}_k, \beta_k)}(\mathbf{X}_k) + \text{Conv}_{g(\mathbf{b}_c, \beta_c)}(\mathbf{X}_c).
    \label{eq:token_generation}
    \vspace{-0.05cm}
\end{equation}

The resulting tokens $\mathbf{T}$ are then processed by a multi-layer Transformer encoder to generate the backbone feature representation $\mathbf{D} \in \mathbb{R}^{\frac{h}{p} \times \frac{w}{p} \times j}$, matching the spatial and channel dimensions of the input tokens $\mathbf{T}$.

As shown in Figure~\ref{fig:hsis_processing}(c), we leverage the learned feature $\mathbf{D}$ as the encoder output for downstream tasks in HyperVision. For different application scenarios, we only need to train a lightweight task model with minimal parameters, which enables efficient adaptation across diverse hyperspectral imaging tasks.

\vspace{-0.45cm}
\section{Unsupervised Representation Learning}
\vspace{-0.25cm}

\begin{figure}[h!]
    \centering
    \vspace{-0.6cm}
    \includegraphics[width=0.7\linewidth]{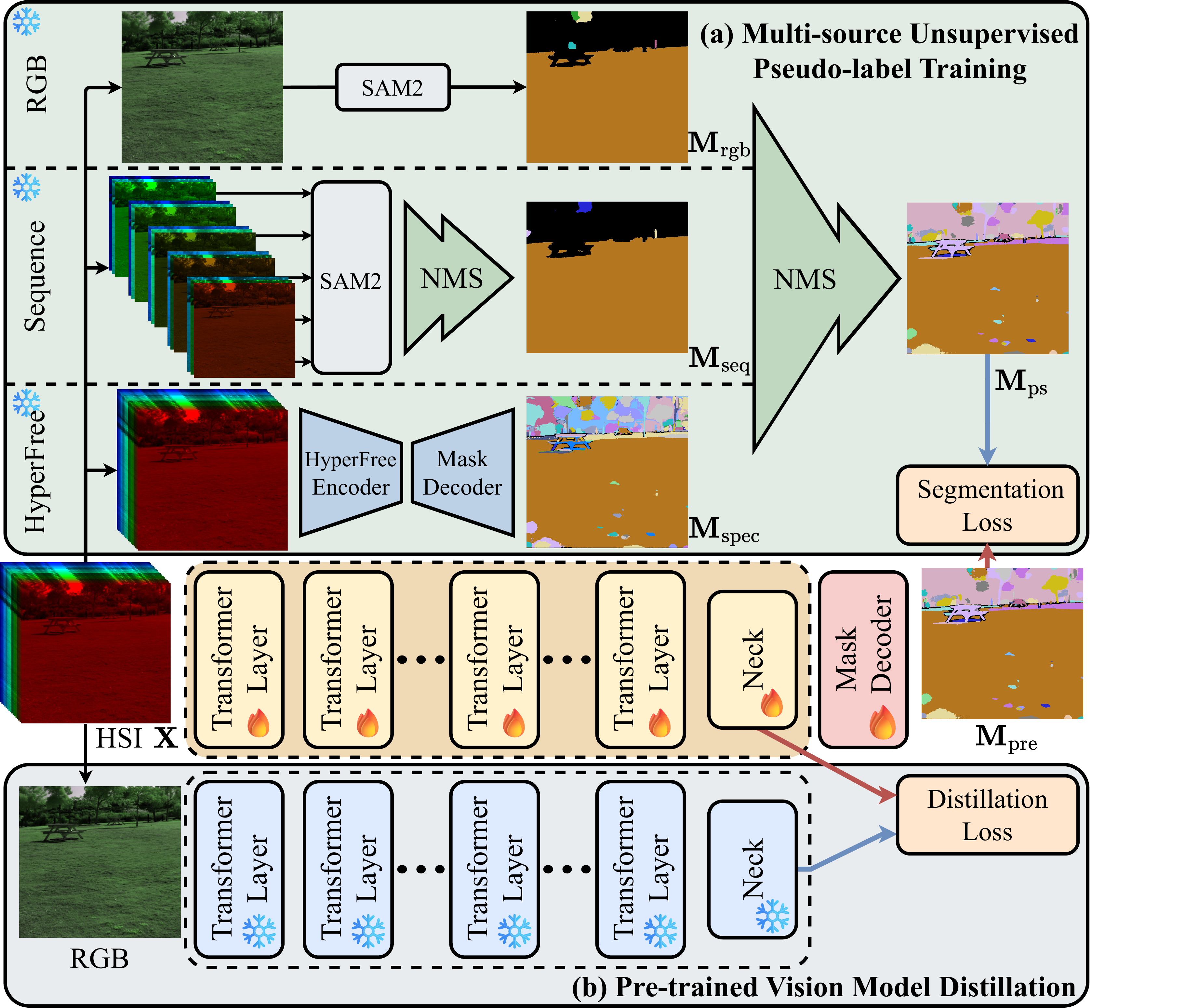}
    \vspace{-0.2cm}
    \caption{Unsupervised representation learning with HyperVision.}
    \vspace{-0.5cm}
    \label{fig:training}
\end{figure}

To address the critical challenges of limited annotations, incompatible labeling schemes, and limited scene diversity in ground-based hyperspectral data, we assemble a comprehensive pre-training collection of approximately 15k images from 26 datasets, and train the network using unsupervised representation learning based on two components: multi-source pseudo-label generation, as shown in Figure~\ref{fig:training}(a), and pre-trained vision model distillation, in Figure~\ref{fig:training}(b).

\vspace{-0.3cm}
\subsection{Ground-based Hyperspectral Images}
\vspace{-0.15cm}

We assembled a large-scale collection from 26 publicly available datasets, comprising 31 subset entries, spanning 377-1,700 nm wavelengths and containing 14,959 diverse images, as shown in Table~\ref{tab:datasets}. The spectral bands range from 15 to 420. For the HOT-2024 \cite{hsotracking} video dataset, we sampled one out of every 100 frames to avoid excessive scene redundancy. This assembled collection provides comprehensive and heterogeneous spectral information for robust pre-training.

\begin{table*}[h!]
    \centering
    \vspace{-0.25cm}
    \caption{Summary of Hyperspectral Datasets}
    \label{tab:datasets}
    \scriptsize
    \resizebox{\linewidth}{!}{
        \begin{tabular}{@{}lclc | lclc@{}}
            \toprule
            \multicolumn{1}{c}{Dataset Name}   & \multicolumn{1}{c}{\#Bands} & \multicolumn{1}{c}{Wavelengths} & \multicolumn{1}{c|}{\#Images} & \multicolumn{1}{c}{Dataset Name}                                    & \multicolumn{1}{c}{\#Bands} & \multicolumn{1}{c}{Wavelengths} & \multicolumn{1}{c}{\#Images} \\
            \midrule
            50 Outdoor \cite{50outdoor}        & 33                          & 400-720 nm                      & 50                            & HyKo v2-NIR \cite{hyko}                                             & 25                          & 600-975 nm                      & 78                           \\
            Agricultural Plant \cite{aphid}    & 237                         & 436-965 nm                      & 361                           & HyKo v2-VIS \cite{hyko}                                             & 16                          & 470-630 nm                      & 163                          \\
            ARAD1K16 \cite{arad1k16}           & 16                          & 400-1000 nm                     & 950                           & HyperBlood \cite{hyperblood}                                        & 128                         & 377-1046 nm                     & 14                           \\
            ARAD1K31 \cite{arad1k31}           & 31                          & 400-700 nm                      & 949                           & HyperDrive-VNIR \cite{hyperdrive}                                   & 24                          & 660-900 nm                      & 504                          \\
            CAVE \cite{cave}                   & 31                          & 400-700 nm                      & 32                            & HyperspectralCity v2 \cite{hyperspectralcityv1,hyperspectralcityv2} & 128                         & 450-950 nm                      & 1330                         \\
            DeepHS-NIR \cite{deephs}           & 252                         & 950-1700 nm                     & 718                           & ICVL \cite{icvl}                                                    & 31                          & 400-700 nm                      & 187                          \\
            DeepHS-VIS \cite{deephs}           & 224                         & 400-1000 nm                     & 3405                          & LIB-HSI \cite{libhsi}                                               & 204                         & 400-1000 nm                     & 393                          \\
            DeepHS-VISCOR \cite{deephs}        & 249                         & 400-1000 nm                     & 1566                          & UM-EMM \cite{um}                                                    & 33                          & 400-720 nm                      & 3                            \\
            Harvard \cite{harvard}             & 31                          & 420-720 nm                      & 77                            & UM-LD 2015 \cite{um}                                                & 33                          & 400-720 nm                      & 20                           \\
            HOT-2024-NIR \cite{hsotracking}    & 25                          & 665-960 nm                      & 477                           & UM-NS 2002 \cite{um}                                                & 31                          & 410-710 nm                      & 8                            \\
            HOT-2024-RedNIR \cite{hsotracking} & 15                          & 600-850 nm                      & 348                           & UM-NS 2004 \cite{um}                                                & 33                          & 400-720 nm                      & 10                           \\
            HOT-2024-VIS \cite{hsotracking}    & 16                          & 470-600 nm                      & 1070                          & UM-OS \cite{um}                                                     & 33                          & 400-720 nm                      & 50                           \\
            HSI Drive v2.0 \cite{hsidrive20}   & 25                          & 600-975 nm                      & 752                           & UM-RI 2015 \cite{um}                                                & 33                          & 400-720 nm                      & 33                           \\
            HSI Road \cite{hsiroad}            & 25                          & 600-960 nm                      & 380                           & Virginia Tech Tree \cite{virginiatree}                              & 420                         & 400-1000 nm                     & 51                           \\
            HSODBIT v2 \cite{hsodbitv2}        & 200                         & 400-1000 nm                     & 500                           & Apple Fire Blight \cite{VNIHDHIATLIMAFB}                            & 204                         & 400-1000 nm                     & 420                          \\
            HSSOD \cite{HSSOD}                 & 81                          & 380-720 nm                      & 60                            & Sum of All Datasets                                                 &                             &                                 & 14959                        \\
            \bottomrule
        \end{tabular}
    }
\end{table*}

\vspace{-0.3cm}
\subsection{Multi-source Pseudo-Labeling}
\vspace{-0.1cm}

\begin{figure}[ht]
    \centering
    \vspace{-0.2cm}
    \includegraphics[width=\linewidth]{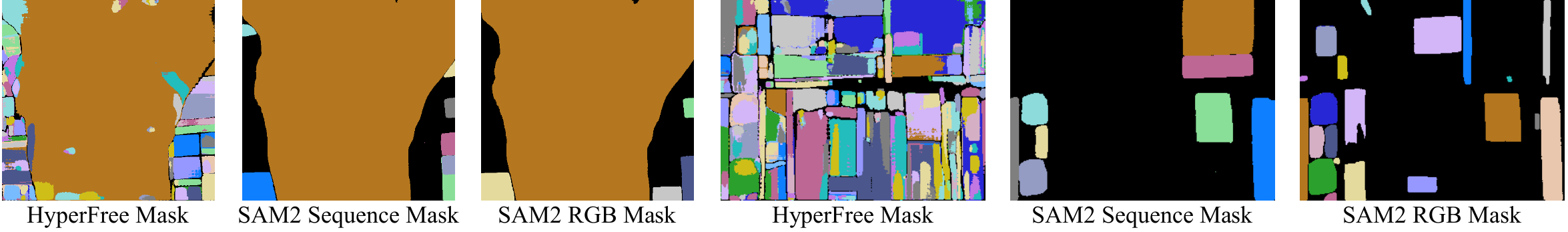}
    \vspace{-0.7cm}
    \caption{Visualization of pseudo-masks generated by SAM2 and HyperFree.}
    \vspace{-0.45cm}
    \label{fig:masks_comparison}
\end{figure}

Learning a unified representation for ground-based hyperspectral data is challenged by limited annotations and incompatible labeling schemes across disparate datasets and downstream tasks. To bypass these constraints, we propose a multi-source pseudo-label generation strategy that leverages pre-trained vision models to automatically produce pseudo-masks $\mathbf{M}_\text{ps}$. As illustrated in Figure~\ref{fig:masks_comparison}, relying on a single source of knowledge is often insufficient: SAM2 provides reliable spatial structures and object boundaries but lacks fine-grained material discrimination, while HyperFree captures material-level semantics from subtle spectral variations but may lack the capability to learn holistic object-level representations. Therefore, as illustrated in Figure~\ref{fig:training}(a), our framework generates candidate masks from three complementary branches to capture comprehensive spatial and spectral semantics, and fuses them to combine the strengths of both sources.

\vspace{-0.3cm}
\subsubsection{RGB-based Spatial Masking:}
\vspace{-0.1cm}
Pre-trained vision models like SAM2~\cite{sam2} excel at zero-shot spatial reasoning to understand complex spatial structures and extract high-level semantics. We leverage these priors by mapping the hyperspectral input to a 3-channel RGB image $\mathbf{I}_\text{RGB}$, either using native RGB alignments or synthesized false-color images. We then employ SAM2 to predict initial spatial masks $\mathbf{M}_\text{RGB} = \text{SAM2}(\mathbf{I}_\text{RGB})$, providing high-quality structural boundaries.

\vspace{-0.3cm}
\subsubsection{Sequential Spatial-Spectral Masking:}
\vspace{-0.1cm}

To capture consistent structural features across the spectral range, we partition the hyperspectral cube into a set of false-color images $\mathcal{I} = \{\mathbf{I}_1, \dots, \mathbf{I}_N\}$, where $N = \frac{\text{channel}}{3}$. Each image $\mathbf{I}_t$ is constructed by sampling three bands with a fixed spectral stride. We leverage SAM2's memory mechanism to maintain spatial coherence across the spectrum:
\begin{equation}
    \mathbf{M}_t = \text{SAM2}(\mathbf{I}_t, \mathcal{M}), \quad t \in \{1, \dots, N\}
\end{equation}
where $\mathcal{M}$ denotes the memory bank that aggregates structural cues from other spectral segments, conditioning the segmentation of the current image $\mathbf{I}_t$. Finally, the resulting mask sequence $\{\mathbf{M}_t\}_{t=1}^N$ is consolidated via intra-branch Non-Maximum Suppression (NMS) to suppress band-specific artifacts and noise, yielding a robust sequence mask $\mathbf{M}_\text{seq}$.

\vspace{-0.3cm}
\subsubsection{HyperFree-based Material Masking:}
\vspace{-0.1cm}

While SAM2 effectively models complex spatial structures and extracts general semantics, it lacks the specialized capacity for fine-grained material discrimination. To acquire semantics deeply rooted in subtle spectral variations, we utilize HyperFree~\cite{Li2025hyperfree}, a channel-adaptive pre-trained model designed to perceive hyperspectral semantics. By capturing diagnostic absorption features that define specific material compositions, HyperFree inherently distinguishes targets that may appear identical in the RGB domain but possess distinct physical properties. By processing the full hyperspectral input $\mathbf{X}$ directly with its pre-trained backbone, we obtain the spectral material mask $\mathbf{M}_{\text{spec}}$, which features fine-grained material-level semantic representations, to complement SAM2's structural priors.

\vspace{-0.3cm}
\subsubsection{Mask Fusion via Non-Maximum Suppression (NMS):} \label{sec:mask_fusion}
\vspace{-0.1cm}

To consolidate the diverse predictions from the three branches into a unified label, we implement a global IoU-based NMS strategy. NMS is applied to suppress redundant masks whose IoU with any higher-scored mask exceeds a threshold $\tau$. The final pseudo-mask $\mathbf{M}_\text{ps}$ is then aggregated via a pixel-wise logical OR over the retained instance-level binary masks:
\begin{equation}
    \begin{split}
        \mathcal{M}_\text{NMS} & = \text{NMS}(\{\mathbf{M}_\text{RGB},\, \mathbf{M}_\text{seq},\, \mathbf{M}_\text{spec}\}, \tau) \\
        \mathbf{M}_\text{ps}   & = \bigvee_{m \in \mathcal{M}_\text{NMS}} m.
    \end{split}
    \label{eq:mps}
\end{equation}
where $m$ denotes an instance-level binary mask retained in $\mathcal{M}_\text{NMS}$, and $\bigvee$ denotes the pixel-wise logical OR operation. This fused mask $\mathbf{M}_\text{ps}$ integrates semantic representations from the complex spatial structures of a pre-trained RGB vision model and the fine-grained spectral material information of a hyperspectral model, providing a comprehensive pseudo-target for optimizing the HyperVision backbone.

\begin{figure}[h]
    \centering
    \vspace{-0.2cm}
    \includegraphics[width=0.55\textwidth]{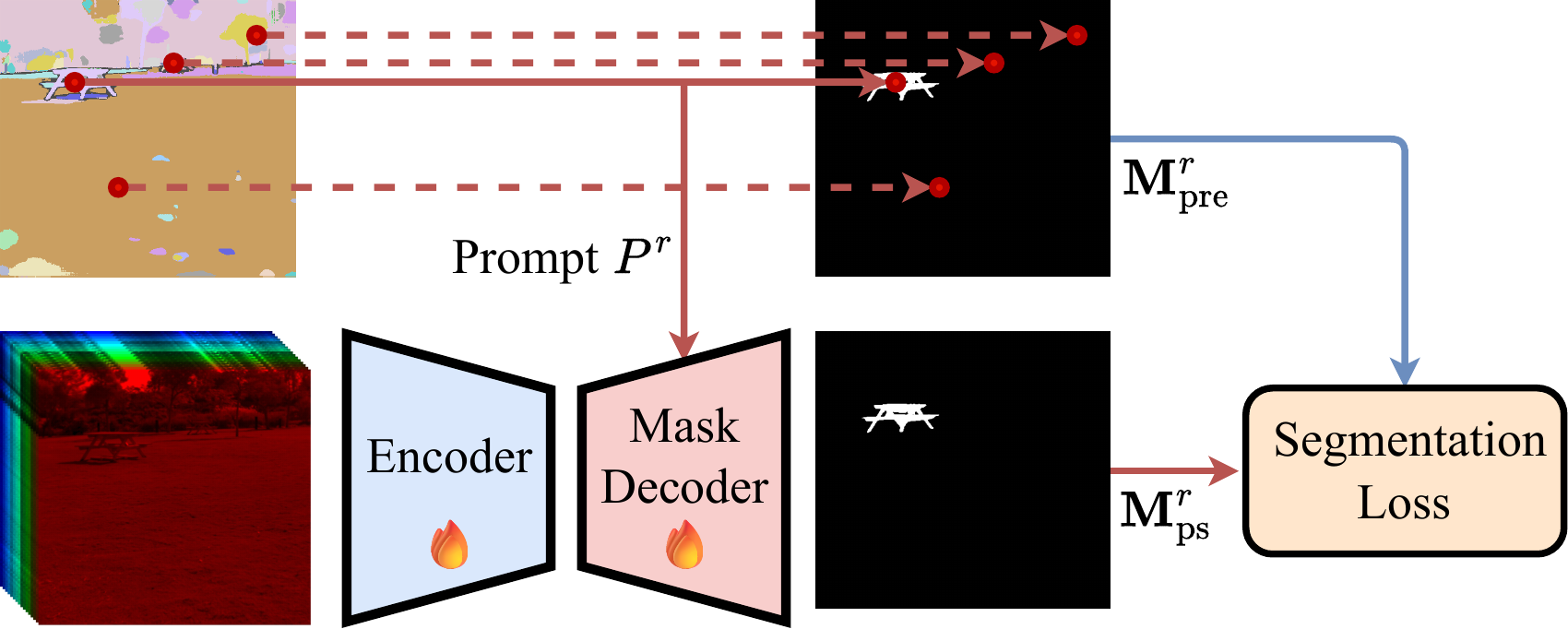}
    \vspace{-0.2cm}
    \caption{Illustration of the prompt-driven segmentation pipeline.}
    \vspace{-0.5cm}
    \label{fig:mask_learn}
\end{figure}

\vspace{-0.3cm}
\subsubsection{Prompt-Driven Segmentation Objective:} \label{sec:seg_objective}
\vspace{-0.1cm}

We adopt a prompt-driven training paradigm, as shown in Figure~\ref{fig:mask_learn}, where the consolidated pseudo-target $\mathbf{M}_\text{ps}$ is decomposed into a set of $R$ distinct sub-masks $\{\mathbf{M}_\text{ps}^r\}_{r=1}^R$. For each sub-mask, a corresponding prompt $\mathbf{P}^r$ and the encoder features $\mathbf{F}$ are fed into a mask decoder to predict the target mask $\mathbf{M}_\text{pre}^r$, where $\mathbf{F} = \text{Neck}(\mathbf{D})$ and $\mathbf{D}$ is the backbone feature:
\begin{equation}
    \begin{split}
        \mathbf{M}_\text{pre}^r & = \text{MaskDecoder}(\mathbf{F}, \mathbf{P}^r).
    \end{split}
\end{equation}

The segmentation loss $L_\text{seg}$ minimizes the discrepancy across all $R$ masks using focal loss, dice loss, and mean squared error. Through this pseudo-label supervision, the HyperVision backbone learns robust structural representations and material-aware features from diverse hyperspectral scenes without any manual labels, forming the core of our unsupervised representation learning framework for downstream perception tasks.

\vspace{-0.3cm}
\subsection{Pre-trained Vision Model Distillation}
\vspace{-0.1cm}

While multi-source pseudo-labeling provides robust supervision, ground-based hyperspectral data inherently suffers from limited scene diversity compared to massive RGB vision datasets. To address this, we introduce a cross-modal pre-trained vision model distillation framework. By employing a frozen, large-scale pre-trained RGB vision model as a teacher, we distill its robust target-level priors into the hyperspectral student network, significantly enhancing the generalization capacity of HyperVision.

\vspace{-0.3cm}
\subsubsection{DINO-style Feature Distillation:}
\vspace{-0.1cm}
As shown in Figure~\ref{fig:training}(b), to achieve efficient knowledge transfer, we leverage the pre-trained Segment Anything Model (SAM)~\cite{Kirillov_2023_ICCV} as the teacher network. Following the distillation paradigm of DINO~\cite{simeoni2025dinov3}, we employ a ``last-stage'' strategy aligning high-level semantic representations. Specifically, we extract the projected feature $\mathbf{F}_S$ from the student network and the frozen feature $\mathbf{F}_T$ from the SAM teacher. The distillation loss $L_\text{dis}$ minimizes the cross-entropy between their spatial token distributions:
\begin{equation}
    L_\text{dis} = -\frac{1}{N_{tok}} \sum_{i=1}^{N_{tok}} \text{Softmax}\left(\mathbf{F}_T^{i}\right) \log \left( \text{Softmax}\left(\text{Linear}(\mathbf{F}_S^{i})\right) \right),
\end{equation}
where $N_{tok}$ denotes the total number of spatial tokens in the feature map. This guides the student to internalize robust visual priors from the RGB vision domain, substantially improving the generalization capacity of HyperVision.

\vspace{-0.3cm}
\subsubsection{Optimization and Adaptation:}
\vspace{-0.1cm}
The joint training objective combines the segmentation and distillation losses:
\begin{equation}
    L_{total} = L_{seg} + \lambda_{dis} L_{dis}.
\end{equation}
This optimization yields a universal frozen pre-trained backbone with transferable feature representations. For downstream tasks, as shown in Figure~\ref{fig:hsis_processing}(c), we perform efficient head-only adaptation. This parameter-efficient strategy limits the need for full network fine-tuning, reduces overall computational overhead, and mitigates overfitting on small HSI datasets, demonstrating practical cross-task flexibility while maintaining competitive performance.

\vspace{-0.4cm}
\section{Experiments} \label{sec:experiments}
\vspace{-0.2cm}

We evaluate the proposed HyperVision framework on three downstream hyperspectral tasks: \textbf{semantic segmentation}, \textbf{object tracking}, and \textbf{salient object detection}. For all tasks, the pre-trained HyperVision backbone remains strictly frozen, and only lightweight task-specific heads are fine-tuned. To highlight the benefits of ground-based pre-training, we also compare our model against airborne pre-trained HyperFree~\cite{Li2025hyperfree} by evaluating it under the exact same frozen backbone protocol and lightweight decoder architectures for a fair comparison. We present quantitative results and visual comparisons for each task, followed by ablation studies to analyze the contribution of the core components.

\vspace{-0.3cm}
\subsection{Hyperspectral Semantic Segmentation} \label{subsec:exp_seg}
\vspace{-0.1cm}

\subsubsection{Datasets and Baselines:}
\vspace{-0.15cm}
We evaluate spatial-spectral perception on the hyperspectral semantic segmentation task using two benchmark datasets: 1) \textbf{HSI Drive v2.0}~\cite{hsidrive20} covering 600-975 nm with 25 bands across 752 images; and 2) \textbf{Hyperspectral City v2}~\cite{hyperspectralcityv1, hyperspectralcityv2} covering 450-950 nm with 128 bands across 1,330 images. We compare HyperVision against three hyperspectral semantic segmentation methods, U-Net~\cite{unet}, DeepLabV3+~\cite{deeplabv3}, and RU-Net~\cite{hs3bench}, and a recently proposed RGB-based method CGRSeg~\cite{cgrseg}. HyperVision and HyperFree~\cite{Li2025hyperfree} adopt the same segmentation decoder architecture as RU-Net, with the backbone frozen during adaptation. CGRSeg is trained and evaluated with the aligned RGB images included in each dataset.
\begin{table*}[htbp]
    \setlength{\tabcolsep}{4pt}
    \centering
    \caption{Comparison of Different Methods for Hyperspectral Semantic Segmentation.}
    \resizebox{\linewidth}{!}{
        \begin{tabular}{llccccc}
            \toprule
            Dataset & Method                                                                & \scalebox{0.82}[1]{Trainable Params} & $\mathrm{Acc}_{\mu}\uparrow$ & $\mathrm{Acc}_{\mathrm{M}}\uparrow$ & $\mathrm{F1}_{\mathrm{M}}\uparrow$ & $\mathrm{J}_{\mathrm{M}}\uparrow$ \\
            \midrule
            \multirow{6}{*}{HSI Drive v2.0 \cite{hsidrive20}}
                    & CGRSeg \cite{cgrseg} (False-color)                                    & 19.1 M                               & 81.02                        & 48.19                               & 58.06                              & 40.96                             \\
            \cline{2-7}
                    & U-Net  \cite{unet}                                                    & 17.3 M                               & 94.74                        & 75.27                               & 76.70                              & 65.37                             \\
                    & \scalebox{1}[1]{DeepLabV3+ (MobileNet) \cite{deeplabv3, mobilenetv2}} & \textbf{5.2 M}                       & 93.79                        & 73.16                               & 75.50                              & 63.57                             \\
                    & \scalebox{1}[1]{DeepLabV3+ (ResNet101) \cite{deeplabv3, resnet} }     & 58.8 M                               & 94.23                        & 73.91                               & 74.98                              & 63.46                             \\
                    & RU-Net \cite{hs3bench}                                                & 21.1 M                               & 95.53                        & 77.09                               & 79.36                              & 68.61                             \\
                    & HyperFree \cite{Li2025hyperfree}                                      & 15.7 M                               & 96.56                        & 87.52                               & 88.22                              & 79.75                             \\
                    & HyperVision                                                           & 15.7 M                               & \textbf{97.44}               & \textbf{89.67}                      & \textbf{90.70}                     & \textbf{83.51}                    \\
            \midrule
            \multirow{6}{*}{\scalebox{0.82}[1]{Hyperspectral City v2 \cite{hyperspectralcityv1, hyperspectralcityv2}}}
                    & CGRSeg \cite{cgrseg} (False-color)                                    & 19.1 M                               & 89.56                        & 56.31                               & 63.46                              & 47.07                             \\
            \cline{2-7}
                    & U-Net    \cite{unet}                                                  & 17.3 M                               & 88.95                        & 54.25                               & 56.28                              & 45.63                             \\
                    & \scalebox{1}[1]{DeepLabV3+(MobileNet) \cite{deeplabv3, mobilenetv2}}  & \textbf{5.2 M}                       & 91.49                        & 56.56                               & 60.13                              & 49.71                             \\
                    & \scalebox{1}[1]{DeepLabV3+(ResNet101) \cite{deeplabv3, resnet}}       & 58.8 M                               & 91.76                        & 58.57                               & 62.23                              & 51.85                             \\
                    & RU-Net \cite{hs3bench}                                                & 21.2 M                               & 93.30                        & 63.13                               & 66.45                              & 56.11                             \\
                    & HyperFree \cite{Li2025hyperfree}                                      & 15.9 M                               & 90.14                        & 58.88                               & 60.56                              & 49.07                             \\
                    & HyperVision                                                           & 15.9 M                               & \textbf{93.41}               & \textbf{66.05}                      & \textbf{67.87}                     & \textbf{57.04}                    \\
            \bottomrule
        \end{tabular}
    }
    \vspace{-0.45cm}
    \label{tab:seg_results}
\end{table*}

\begin{figure*}[t]
    \centering
    \vspace{-0.1cm}
    \includegraphics[width=\linewidth]{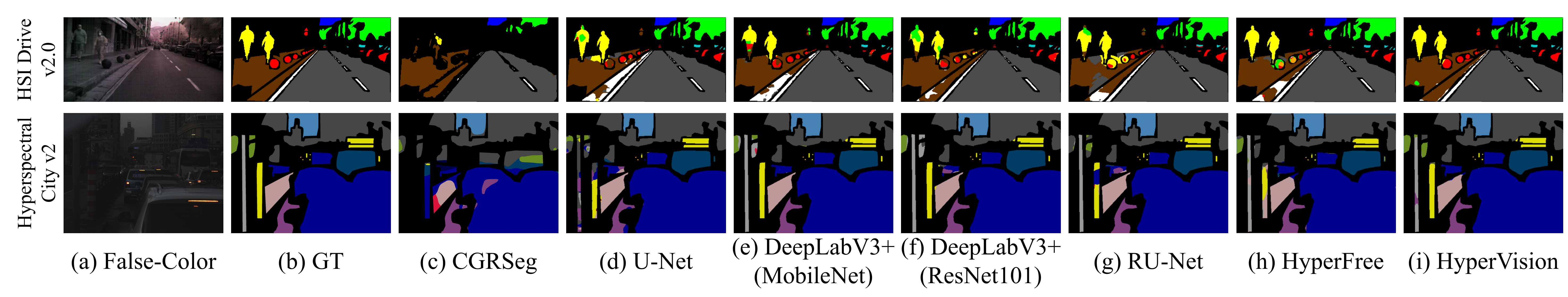}
    \vspace{-0.75cm}
    \caption{Visual comparisons of hyperspectral semantic segmentation results.}
    \vspace{-0.45cm}
    \label{fig:seg_vis}
\end{figure*}

\vspace{-0.3cm}
\subsubsection{Results and Visualization:}
\vspace{-0.1cm}
As shown in Table~\ref{tab:seg_results}, HyperVision achieves state-of-the-art performance across all metrics on both datasets. With a head-only adaptation setup, our method outperforms both pre-trained and task-specific baselines, yielding up to a 16.3\% relative improvement in $\mathrm{Acc}_{\mathrm{M}}$ over the state-of-the-art task-specific method, RU-Net~\cite{hs3bench}. Notably, HyperVision further improves upon HyperFree~\cite{Li2025hyperfree}. The visual comparisons in Figure~\ref{fig:seg_vis} show that HyperVision produces clearer decision boundaries and maintains better semantic integrity compared to the other methods.

\vspace{-0.3cm}
\subsection{Hyperspectral Object Tracking} \label{subsec:exp_tracking}
\vspace{-0.1cm}

\subsubsection{Datasets and Baselines:}
\vspace{-0.15cm}

For temporal and spectral perception, we evaluate on the \textbf{HOT-2023} \cite{hsotracking} benchmark, which is more widely adopted for evaluation in existing tracking applications than HOT-2024. It contains 16 sequences with NIR, RedNIR, and VIS subsets, which span 470-960 nm with 15-25 bands. We compare against 7 state-of-the-art hyperspectral trackers: MHT \cite{mht}, TSCFW \cite{tscfw}, BAE-Net \cite{baenet}, SiamF \cite{siamf}, SiamBAG \cite{siambag}, SEE-Net \cite{Li2023DENet}, and SENSE \cite{sense}. HyperVision adopts the tracking decoder from SiamHYPER~\cite{siamhyper} and fine-tunes it with the backbone frozen.

\vspace{-0.3cm}
\subsubsection{Results and Visualization:}
\vspace{-0.1cm}
Hyperspectral object tracking is currently a highly prominent and active task in ground-based hyperspectral vision, with all compared baselines featuring carefully designed, high-performance architectures. Despite this formidable competition, quantitative results in Table~\ref{tab:tracking_results} show our model achieves the highest AUC and DP, yielding a 2.1\% relative improvement in AUC over SENSE. Utilizing the frozen backbone, the model requires 25.9~M trainable parameters in the tracking head. Furthermore, under the same evaluation protocol, HyperVision achieves better performance than the airborne pre-trained model HyperFree~\cite{Li2025hyperfree}, demonstrating the clear advantage of ground-based pre-training for egocentric tracking scenarios. As shown in Figure~\ref{fig:tracking_vis}, HyperVision exhibits robust target discrimination ability across the VIS, NIR, and RedNIR sensor subsets, accurately tracking targets against complex and cluttered backgrounds under diverse spectral conditions.

\begin{table}[htbp]
    \setlength{\tabcolsep}{2pt}
    \centering
    \vspace{-0.2cm}
    \caption{Comparison of Different Trackers for Hyperspectral Object Tracking.}
    \resizebox{\linewidth}{!}{
        \begin{tabular}{lccccccccc}
            \toprule
            Method           & BAE-Net\cite{baenet} & MHT\cite{mht} & SEE-Net\cite{Li2023DENet} & SiamBAG\cite{siambag} & SiamF\cite{siamf} & TSCFW\cite{tscfw} & SENSE\cite{sense} & HyperFree\cite{Li2025hyperfree} & HyperVision    \\
            \midrule
            AUC $\uparrow$   & 0.507                & 0.453         & 0.519                     & 0.512                 & 0.511             & 0.469             & 0.559             & 0.562                           & \textbf{0.571} \\
            DP $\uparrow$    & 0.777                & 0.717         & 0.763                     & 0.739                 & 0.767             & 0.703             & 0.766             & 0.769                           & \textbf{0.785} \\
            Trainable Params & \textbf{17.7 M}      & -             & 53.9 M                    & 90.44 M               & 54.8 M            & -                 & 55.7 M            & 25.9 M                          & 25.9 M         \\
            \bottomrule
        \end{tabular}
    }
    \vspace{-0.45cm}
    \label{tab:tracking_results}
\end{table}

\begin{figure*}[htbp]
    \centering
    \setlength{\tabcolsep}{1pt}
    \scriptsize
    \begin{tabular}{cccccc}
        \includegraphics[width=0.16\linewidth]{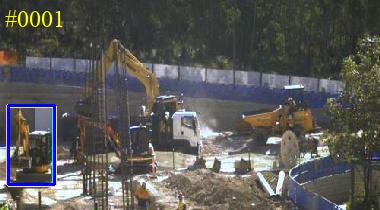}        &
        \includegraphics[width=0.16\linewidth]{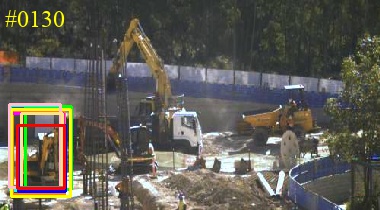}        &
        \includegraphics[width=0.16\linewidth]{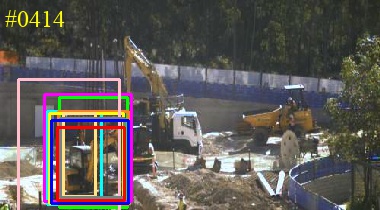}        &
        \includegraphics[width=0.16\linewidth]{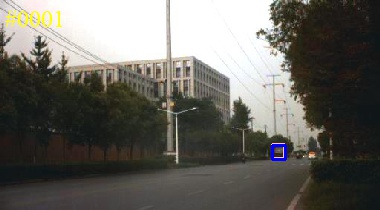}          &
        \includegraphics[width=0.16\linewidth]{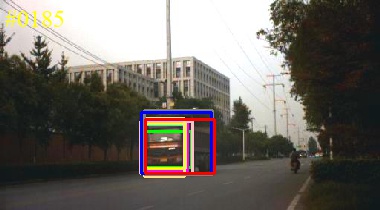}          &
        \includegraphics[width=0.16\linewidth]{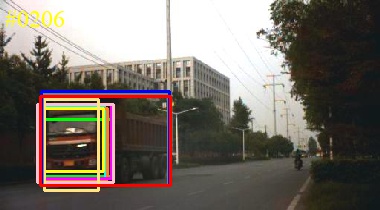}                                                                            \\
        \multicolumn{3}{c}{vis-excavator \#0001, \#0130, \#0414}                          & \multicolumn{3}{c}{vis-trucker \#0001, \#0185, \#0206}          \\
        \includegraphics[width=0.16\linewidth]{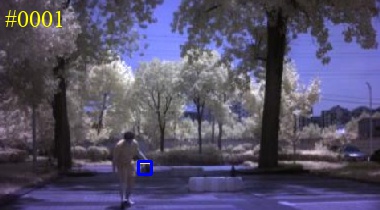}      &
        \includegraphics[width=0.16\linewidth]{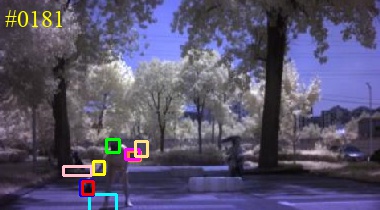}      &
        \includegraphics[width=0.16\linewidth]{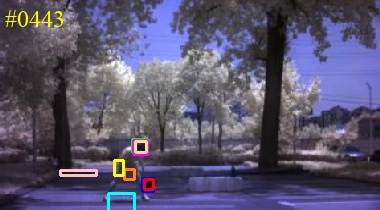}      &
        \includegraphics[width=0.16\linewidth]{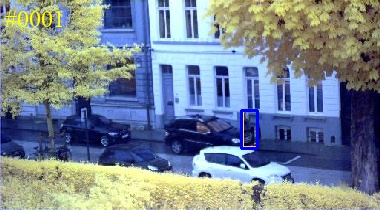} &
        \includegraphics[width=0.16\linewidth]{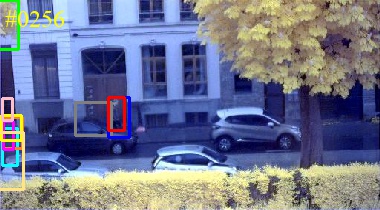} &
        \includegraphics[width=0.16\linewidth]{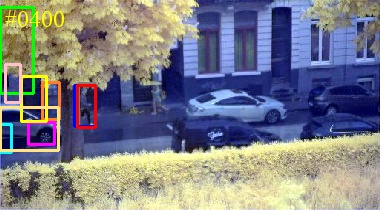}                                                                   \\
        \multicolumn{3}{c}{nir-basketball3 \#0001, \#0181, \#0443}                        & \multicolumn{3}{c}{rednir-rainystreet16 \#0001, \#0256, \#0400} \\
    \end{tabular}

    \includegraphics[width=1\linewidth]{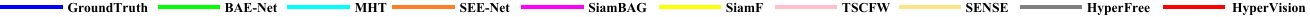}
    \vspace{-0.4cm}
    \caption{Visual comparisons of hyperspectral object tracking results.}
    \vspace{-0.4cm}
    \label{fig:tracking_vis}
\end{figure*}

\vspace{-0.3cm}
\subsection{Hyperspectral Salient Object Detection}

\label{subsec:exp_sod}

\subsubsection{Datasets and Baselines:}
\vspace{-0.1cm}
For this task, we evaluate on the \textbf{HSSOD} dataset \cite{HSSOD}, which contains 60 hyperspectral images with 81 bands spanning 380-720 nm. We compare against hyperspectral salient object detection methods, Spatial-Spectral Mediated Network (SMN) \cite{liu2024smn}, with ResNet, Swin-T, and PVT-v2 backbones, and RGB-based salient object detection methods, U-2-Net \cite{qin2020u2net} and BA-SAM \cite{Song_2024_CVPR}. HyperVision and HyperFree \cite{Li2025hyperfree} adopt the same detection decoder from BA-SAM~\cite{Song_2024_CVPR}, and both of them freeze their respective backbones during the training. U-2-Net and BA-SAM are trained and evaluated using the aligned color images provided by the dataset.

\vspace{-0.3cm}
\subsubsection{Results and Visualization:}
\begin{figure*}[htbp]
    \vspace{-0.2cm}
    \centering
    \includegraphics[width=1\linewidth]{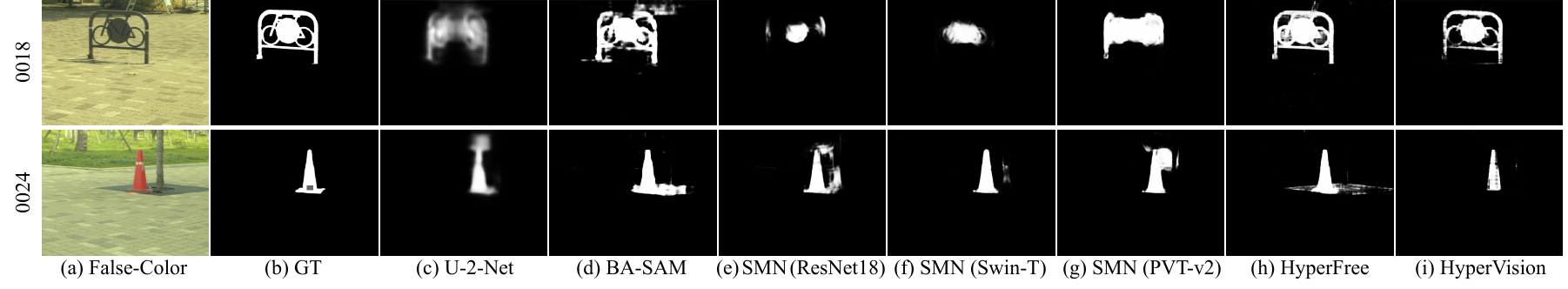}
    \vspace{-0.65cm}
    \caption{Visual comparisons of salient object detection results.}
    \vspace{-0.55cm}
    \label{fig:hssod_vis}
\end{figure*}

\vspace{-0.1cm}
As shown in Table~\ref{tab:performance_params}, HyperVision achieves state-of-the-art performance across all four metrics using head-only adaptation with 4.1~M trainable parameters, yielding a 30\% relative reduction in MAE over the RGB-based method, BA-SAM, and a 35.5\% reduction over the best task-specific hyperspectral method, SMN (Swin-T). Specifically, under the same decoder setup, HyperVision outperforms HyperFree~\cite{Li2025hyperfree} on CC, MAE, and $F_{\beta}^{\mathrm{max}}$. Figure~\ref{fig:hssod_vis} shows that HyperVision provides better background suppression and clearer object boundaries compared to existing approaches.

\begin{table}[htbp]
    \setlength{\tabcolsep}{5pt}
    \centering
    \caption{Comparison of Different Methods for Hyperspectral Salient Object Detection.}
    \vspace{0.1cm}
    \resizebox{0.8\linewidth}{!}{
        \begin{tabular}{lccccc}
            \toprule
            Method                                     & \scalebox{0.82}[1]{Trainable Params} & AUC $\uparrow$ & CC $\uparrow$  & MAE $\downarrow$ & $F_{\beta}^{\mathrm{max}}$ $\uparrow$ \\
            \midrule
            U-2-Net \cite{qin2020u2net} (False-color)  & 44.0 M                               & 0.871          & 0.515          & 0.100            & 0.536                                 \\
            BA-SAM \cite{Song_2024_CVPR} (False-color) & \textbf{4.1 M}                       & 0.938          & 0.665          & 0.070            & 0.681                                 \\
            \hline
            SMN (ResNet18) \cite{liu2024smn, resnet}   & 7.17 M                               & 0.896          & 0.612          & 0.091            & 0.624                                 \\
            SMN (Swin-T) \cite{liu2024smn, swint}      & 16.77 M                              & 0.886          & 0.670          & 0.076            & 0.693                                 \\
            SMN (PVT-v2) \cite{liu2024smn,PVTv2}       & 10.08 M                              & 0.906          & 0.662          & 0.079            & 0.676                                 \\
            HyperFree \cite{Li2025hyperfree}           & \textbf{4.1 M}                       & \textbf{0.959} & 0.735          & 0.052            & 0.728                                 \\
            HyperVision                                & \textbf{4.1 M}                       & \textbf{0.959} & \textbf{0.762} & \textbf{0.049}   & \textbf{0.742}                        \\
            \bottomrule
        \end{tabular}
    }
    \vspace{-0.5cm}
    \label{tab:performance_params}
\end{table}

\vspace{-0.3cm}
\subsection{Ablation Study} \label{subsec:analysis}
\vspace{-0.2cm}

We conduct ablation studies on the HSI Drive v2.0 dataset \cite{hsidrive20} to validate our unsupervised representation learning framework. Specifically, we analyze the scalability of pre-training across different model scales and the individual contributions of the unsupervised components.

\vspace{-0.5cm}
\subsubsection{Effectiveness of Model Scale and Unsupervised Components}
\vspace{-0.15cm}
We evaluate the scalability of our pre-training framework across three model scales (Base, Large, and Huge) under various pre-training configurations in Table~\ref{tab:ablation_scale}, where the first row uses the pre-trained HyperFree~\cite{Li2025hyperfree} as the baseline backbone. This controlled study examines the effects of our pseudo-labeling framework and cross-modal distillation. Evaluating the framework across different model capacities reveals that larger models benefit more from cross-modal distillation, while the Base model shows a slight performance degradation when distillation is applied. We attribute this to the substantially higher information density of HSIs compared to RGB images: the limited model capacity of the Base variant is saturated by the task of extracting fine-grained spectral semantic representations, leaving insufficient capacity to simultaneously internalize the spatial structure priors transferred from the RGB teacher.

\begin{table*}[htbp]
    \centering
    \setlength{\tabcolsep}{3pt}
    \vspace{-0.1cm}
    \caption{Ablation Study of HyperVision Pre-training Components across Different Model Scales on HSI Drive v2.0.}
    \label{tab:ablation_scale}
    \vspace{0.15cm}
    \resizebox{\linewidth}{!}{
        \begin{tabular}{cc cccc cccc cccc}
            \toprule
            \multirow{2}{*}{Pseudo-masks} & \multirow{2}{*}{Distillation} & \multicolumn{4}{c}{Base {\scriptsize(5.7M/150M)}} & \multicolumn{4}{c}{Large {\scriptsize(10.1M/387M)}} & \multicolumn{4}{c}{Huge {\scriptsize(15.7M/751M)}}                                                                                                                                                                                                                                                                                                                                 \\
            \cmidrule(lr){3-6} \cmidrule(lr){7-10} \cmidrule(lr){11-14}
                                          &                               & $\mathrm{Acc}_{\mu}\uparrow$                      & $\mathrm{Acc}_{\mathrm{M}}\uparrow$                 & $\mathrm{F1}_{\mathrm{M}}\uparrow$                 & $\mathrm{J}_{\mathrm{M}}\uparrow$ & $\mathrm{Acc}_{\mu}\uparrow$ & $\mathrm{Acc}_{\mathrm{M}}\uparrow$ & $\mathrm{F1}_{\mathrm{M}}\uparrow$ & $\mathrm{J}_{\mathrm{M}}\uparrow$ & $\mathrm{Acc}_{\mu}\uparrow$ & $\mathrm{Acc}_{\mathrm{M}}\uparrow$ & $\mathrm{F1}_{\mathrm{M}}\uparrow$ & $\mathrm{J}_{\mathrm{M}}\uparrow$ \\
            \midrule
                                          &                               & 95.64                                             & 84.27                                               & 84.89                                              & 75.00                             & 96.21                        & 86.52                               & 87.38                              & 78.41                             & 96.56                        & 87.52                               & 88.22                              & 79.75                             \\
            \checkmark                    &                               & \textbf{96.77}                                    & \textbf{87.42}                                      & \textbf{87.63}                                     & \textbf{79.02}                    & 96.37                        & 85.52                               & 85.57                              & 76.12                             & 97.20                        & 89.49                               & 89.68                              & 81.96                             \\
            \checkmark                    & \checkmark                    & 96.69                                             & 86.63                                               & 87.09                                              & 78.21                             & \textbf{97.04}               & \textbf{88.49}                      & \textbf{88.96}                     & \textbf{80.87}                    & \textbf{97.44}               & \textbf{89.67}                      & \textbf{90.70}                     & \textbf{83.51}                    \\
            \bottomrule
        \end{tabular}
    }
    \vspace{-0.25cm}
\end{table*}

Furthermore, we conduct a comprehensive ablation study to evaluate the individual and combined contributions of the pseudo-mask sources (RGB-based spatial masking $\mathbf{M}_\text{RGB}$, sequential spatial-spectral masking $\mathbf{M}_\text{seq}$, and HyperFree material masking $\mathbf{M}_\text{spec}$) on the Huge model variant, as summarized in Table~\ref{tab:ablation_component}. The row marked with $\ast$ directly employs HyperFree as the frozen backbone without our representation learning. The results demonstrate that $\mathbf{M}_\text{seq}$ achieves better performance than $\mathbf{M}_\text{RGB}$ across all metrics, confirming that sequential spatial-spectral masks provide more reliable structural supervision across band transitions. Moreover, integrating $\mathbf{M}_\text{spec}$ provides crucial material-level semantics that are absent in RGB-based priors. Fusing all three sources yields the best performance across all metrics, demonstrating the complementarity among spatial structure priors, spectral sequence coherence, and fine-grained material cues.

\begin{table}[htbp]
    \centering
    \setlength{\tabcolsep}{4pt}
    \vspace{-0.1cm}
    \caption{Ablation of Pseudo-mask Sources on HSI Drive v2.0.}
    \vspace{0.15cm}
    \label{tab:ablation_component}
    \resizebox{0.65\linewidth}{!}{
        \begin{tabular}{ccc cccc}
            \toprule
            RGB        & Sequence   & HyperFree  & $\mathrm{Acc}_{\mu}\uparrow$ & $\mathrm{Acc}_{\mathrm{M}}\uparrow$ & $\mathrm{F1}_{\mathrm{M}}\uparrow$ & $\mathrm{J}_{\mathrm{M}}\uparrow$ \\
            \midrule
            \checkmark &            &            & 96.38                        & 84.91                               & 86.00                              & 76.58                             \\
                       & \checkmark &            & 96.81                        & 87.98                               & 88.72                              & 80.50                             \\
                       &            & $\ast$     & 96.56                        & 87.52                               & 88.22                              & 79.75                             \\
            \checkmark & \checkmark & \checkmark & \textbf{97.44}               & \textbf{89.67}                      & \textbf{90.70}                     & \textbf{83.51}                    \\
            \bottomrule
        \end{tabular}
    }
    \vspace{-0.35cm}
\end{table}

\vspace{-0.3cm}
\subsection{Computational Cost} \label{subsec:cost}
\vspace{-0.15cm}

We analyze the computational and parameter efficiency of HyperVision alongside task-specific baselines on HSI Drive v2.0 in Table~\ref{tab:cost}. It is important to distinguish parameter-efficient adaptation from computational inference efficiency. As demonstrated across downstream benchmarks (Tables~\ref{tab:seg_results}, \ref{tab:tracking_results}, and \ref{tab:performance_params}), HyperVision enables parameter-efficient adaptation by freezing the backbone and tuning only lightweight task-specific heads (\eg, 4.1~M parameters for salient object detection). Regarding inference cost, while task-specific CNN baselines exhibit lower inference latency, our dense foundation model prioritizes spectral representation completeness and universal cross-sensor generalization over raw speed. In terms of training overhead, pre-training takes approximately 40 hours for each model variant, which remains acceptable given the framework's ability to address heterogeneous spectral configurations and annotation schemes within a unified framework.

\begin{table}[htbp]
    \centering
    \vspace{-0.15cm}
    \caption{Computational Cost Comparison on HSI Drive v2.0.}
    \label{tab:cost}
    \resizebox{\linewidth}{!}{
        \begin{tabular}{l|ccccccc}
            \toprule
            Method    & U-Net   & RU-Net   & DLV3+(MB) & DLV3+(R101) & HV-B     & HV-L      & HV-H      \\
            \midrule
            Inf. Time & 9.25 ms & 13.03 ms & 7.86 ms   & 11.52 ms    & 69.61 ms & 140.43 ms & 221.40 ms \\
            GPU Mem.  & 726 MiB & 714 MiB  & 1100 MiB  & 1382 MiB    & 2904 MiB & 3370 MiB  & 3770 MiB  \\
            \bottomrule
        \end{tabular}}
    \vspace{-0.55cm}
\end{table}

\begin{figure}[htbp]
    \centering
    \vspace{-0.15cm}
    \includegraphics[width=0.76\linewidth]{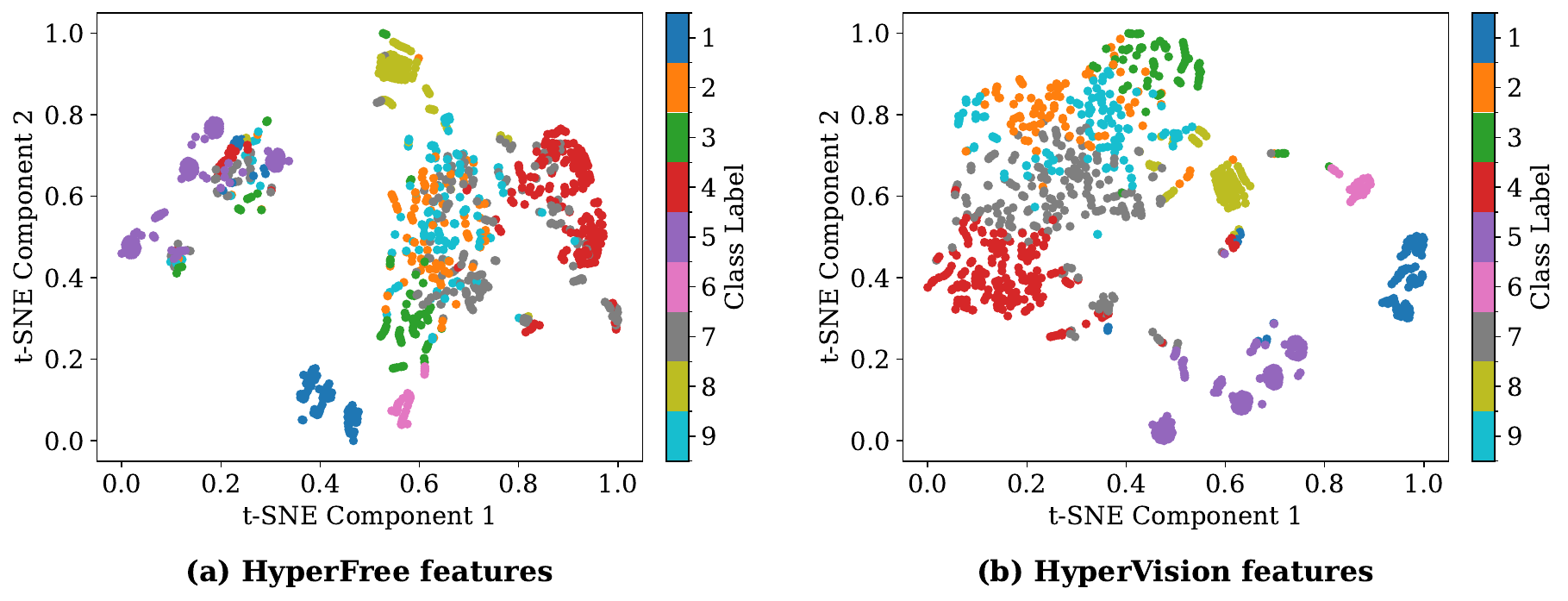}
    \vspace{-0.35cm}
    \caption{t-SNE visualization comparing the feature representations of HyperFree and HyperVision.}
    \vspace{-0.55cm}
    \label{fig:tsne}
\end{figure}

\vspace{-0.2cm}
\subsection{Feature Visualization} \label{subsec:tsne}
\vspace{-0.15cm}

We further provide a t-distributed stochastic neighbor embedding (t-SNE) visualization comparing features from HyperFree \cite{Li2025hyperfree} and the HyperVision model extracted from the HSI Drive v2.0 image \texttt{nf1111\_571}. As shown in Figure~\ref{fig:tsne}, HyperVision exhibits clearer semantic cluster boundaries and tighter intra-class distributions, suggesting it extracts more discriminative target representations.

\vspace{-0.45cm}
\section{Conclusion} \label{sec:conclusion}
\vspace{-0.25cm}
In this paper, we introduced HyperVision, a channel-adaptive pre-trained backbone tailored for ground-based hyperspectral perception. First, to address hardware fragmentation, we adopted a dynamic embedding mechanism that unifies heterogeneous hyperspectral inputs into a shared token space without relying on fixed channel configurations. Second, to address limited annotations and heterogeneous labeling schemes, we designed a multi-source pseudo-label generation strategy that fuses semantic representations from both spatial structures generated by SAM2 and fine-grained spectral material information extracted by HyperFree. Third, to compensate for limited scene diversity, we incorporated a cross-modal knowledge distillation framework that transfers rich visual priors from a large-scale pre-trained RGB vision model to the hyperspectral backbone, enhancing generalization capacity without labeled data. HyperVision is pre-trained on a collection of 26 diverse public datasets comprising 15k HSIs. Extensive experiments across hyperspectral semantic segmentation, object tracking, and salient object detection demonstrate that HyperVision achieves state-of-the-art performance through an efficient, head-only adaptation paradigm. We believe this work provides a robust and scalable foundation for future research in ground-based hyperspectral image understanding.

\section*{Acknowledgements}
This work was supported in part by the Natural Science Foundation of Jiangsu Province (Grant No. BK20250185), the National Natural Science Foundation of China (Grant No. 424B2010), and the State Key Laboratory of Novel Software Technology at Nanjing University (Grant No. KFKT2026B86).
\bibliography{hypervision}

\end{document}